\documentclass[10pt, conference, compsocconf]{IEEEtran}

\ifCLASSINFOpdf
\else
\fi

\usepackage{graphicx}

\begin{document}
%
\title{Fast Glare Detection in Document Images}


\author{\IEEEauthorblockN{Dmitry  Rodin\IEEEauthorrefmark{3}\IEEEauthorrefmark{1}, 
Nikita Orlov\IEEEauthorrefmark{3}\IEEEauthorrefmark{2}}

\IEEEauthorblockA{
    \IEEEauthorrefmark{3}
    Phystech School of Applied Mathematics and Informatics\\
    Moscow Institute of Physics and Technology (National Research University)
}

\IEEEauthorblockA{
    \IEEEauthorrefmark{1}
    R\&D Department\\
    ABBYY Production LLC\\
    Moscow, Russia\\
    d.rodin@abbyy.com 
}

\IEEEauthorblockA{
    \IEEEauthorrefmark{2}
    PicsArt AI\\
    nikita.orlov@picsart.com
}

}



%


\maketitle

\begin{abstract}
Glare is a phenomenon that occurs when the scene has a reflection of a light source or has one in it. This luminescence can hide useful information from the image, making text recognition virtually impossible. In this paper, we propose an approach to detect glare in images taken by users via mobile devices. Our method divides the document into blocks and collects luminance features from the original image and black-white strokes histograms of the binarized image. Finally, glare is detected using a convolutional neural network on the aforementioned histograms and luminance features. The network consists of several feature extraction blocks, one for each type of input, and the detection block, which calculates the resulting glare heatmap based on the output of the extraction part. The proposed solution detects glare with high recall and f-score. 

\end{abstract}


%
\IEEEpeerreviewmaketitle

\section{Introduction}

In the last several years, deep convolutional networks have outperformed human results in visual recognition tasks [1]. However, many tasks demand localization as an output of the network, so a label is assigned to each pixel, or, in some cases, to a region of the image.

Glare is a phenomenon that occurs when the scene has a light source or a reflection of one in it. This luminescence can hide information needed for text detection and recognition, so detecting glare in the image can help in document image quality assessment. Also, glare has high in-class variance: the shape, size, and dynamic range of glare region highly depend on the physical shape of the document and the material. 

Glare detection is an important problem in optical character recognition (OCR) and in document image quality assessment. By locating the glare region we can estimate the recognition results of the document, without running the recognition algorithm. 
If the method is sufficiently fast it can be run on a mobile device in real time. Detection results can increase the text recognition quality capturing a better image without glares.
Several methods have been proposed for image segmentation, but most of them focus on detecting glare in the outdoor environment [2], [3]. Also, the size of the glare region in the document image implies the usage of up to 40\% of the image as the context for detection, which demands great computational resources.

The goal of this paper is to propose a fast and lightweight network, capable of detecting glare in document images on mobile devices in real time. The main goal is to create a detector fast enough to indicate approximate glare region in real time to prevent the user from taking a photo that could not be recognized later. The main idea is to extract luminance features and black-white run length strokes histograms from the image using classic approaches and build a convolutional neural network on these features.

In section II we describe the approach to the data preparation. In section III we describe the network configuration and the approach to model training. In section IV we describe the experimental results.

\section{Data description}
\renewcommand{\figurename}{Fig.}
\begin{figure}
\begin{tabular}{c|c}
\includegraphics[width = 4cm, height=5.2cm]{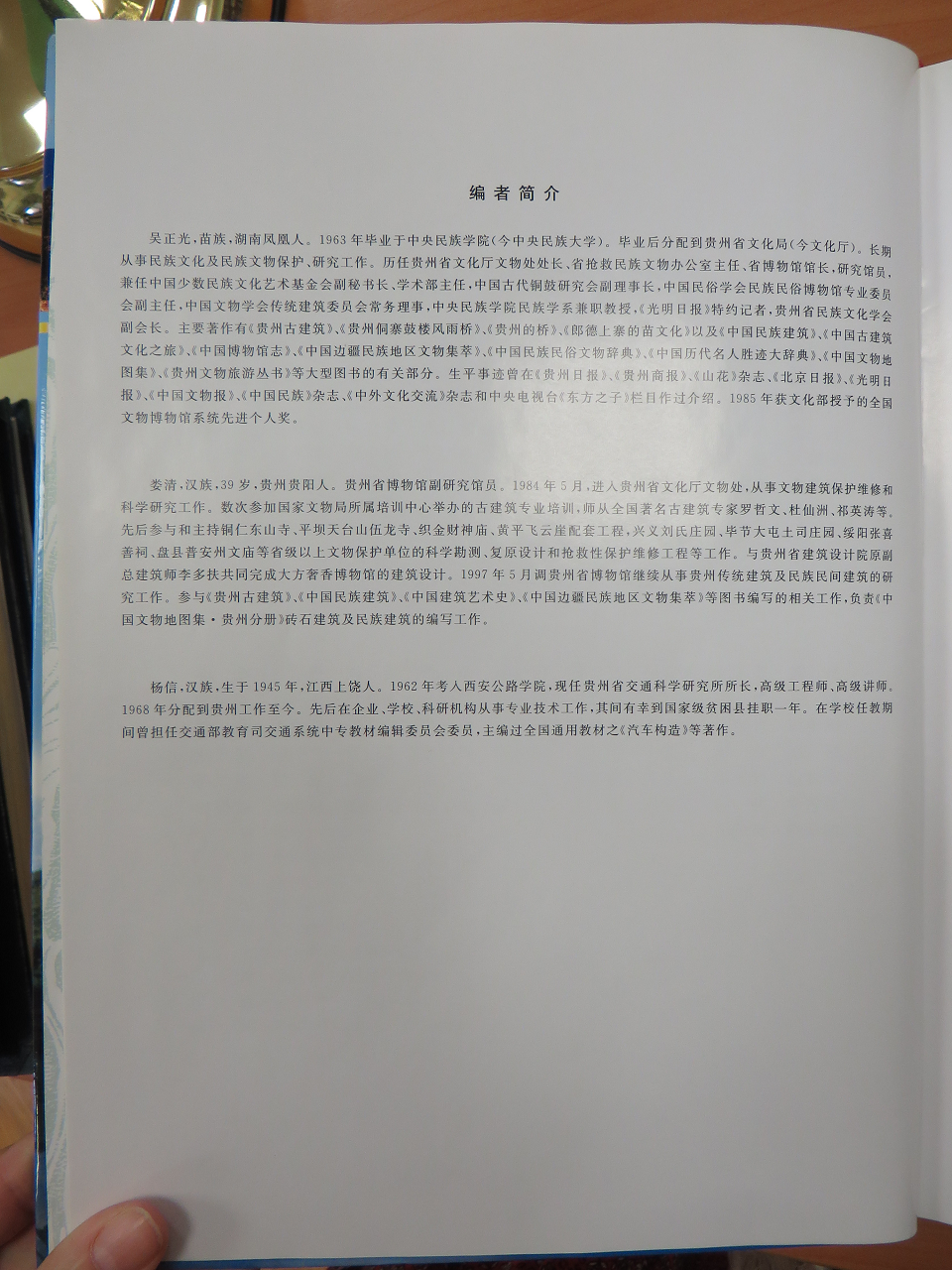}&
\includegraphics[width = 4cm, height=5.2cm]{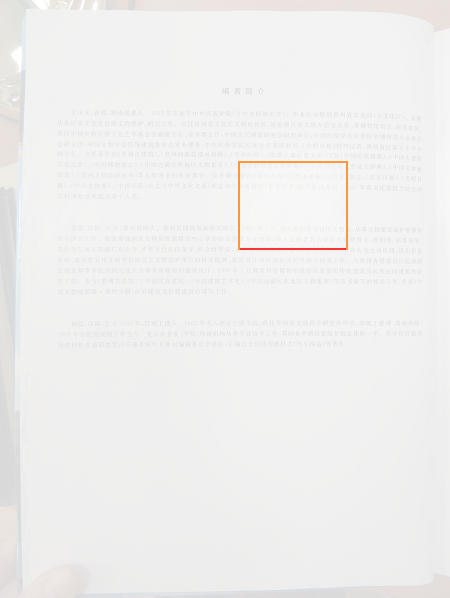}\\
\hline
\includegraphics[width = 4cm, height=5.2cm]{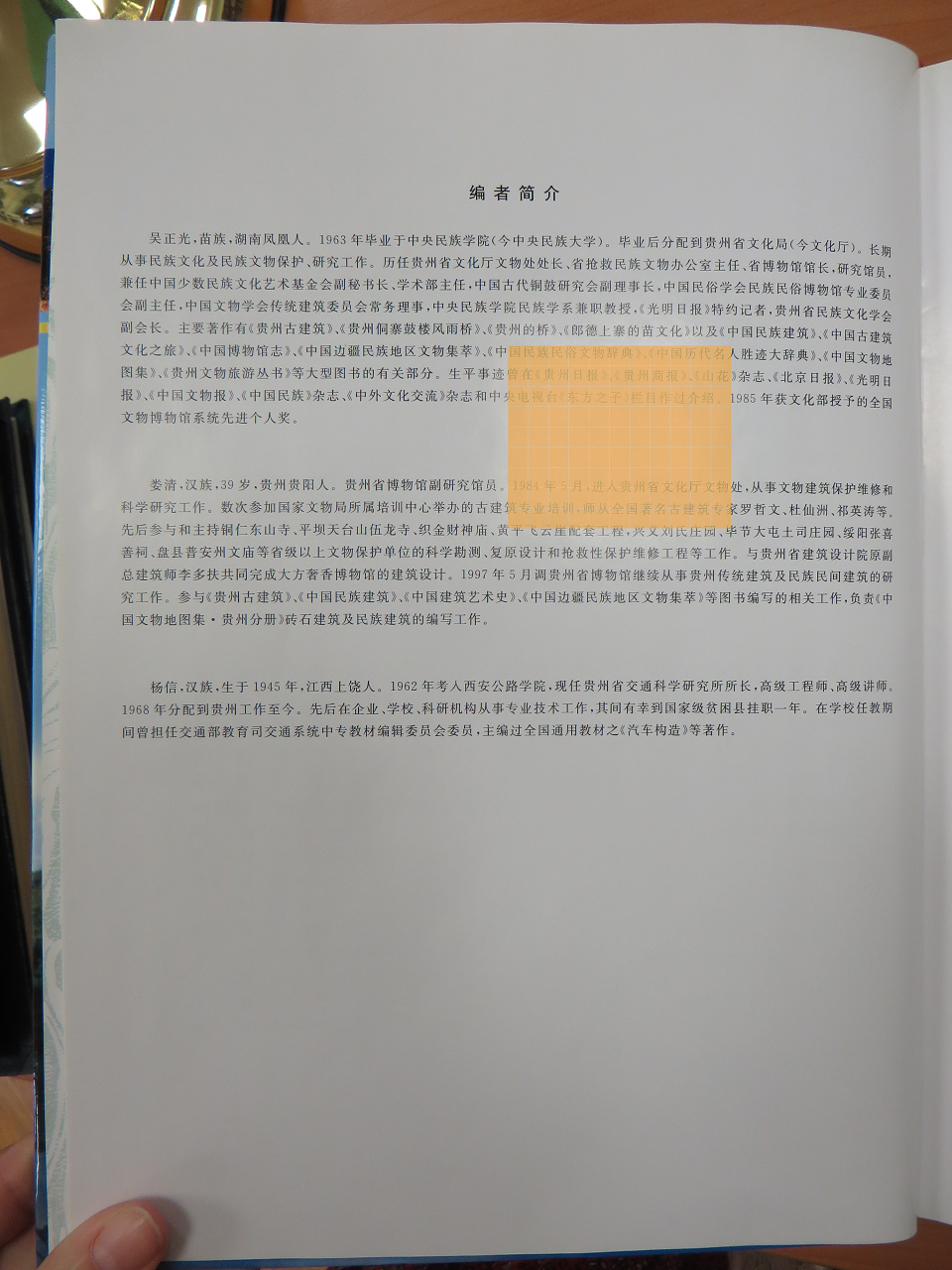} &
\includegraphics[width = 4cm, height=5.2cm]{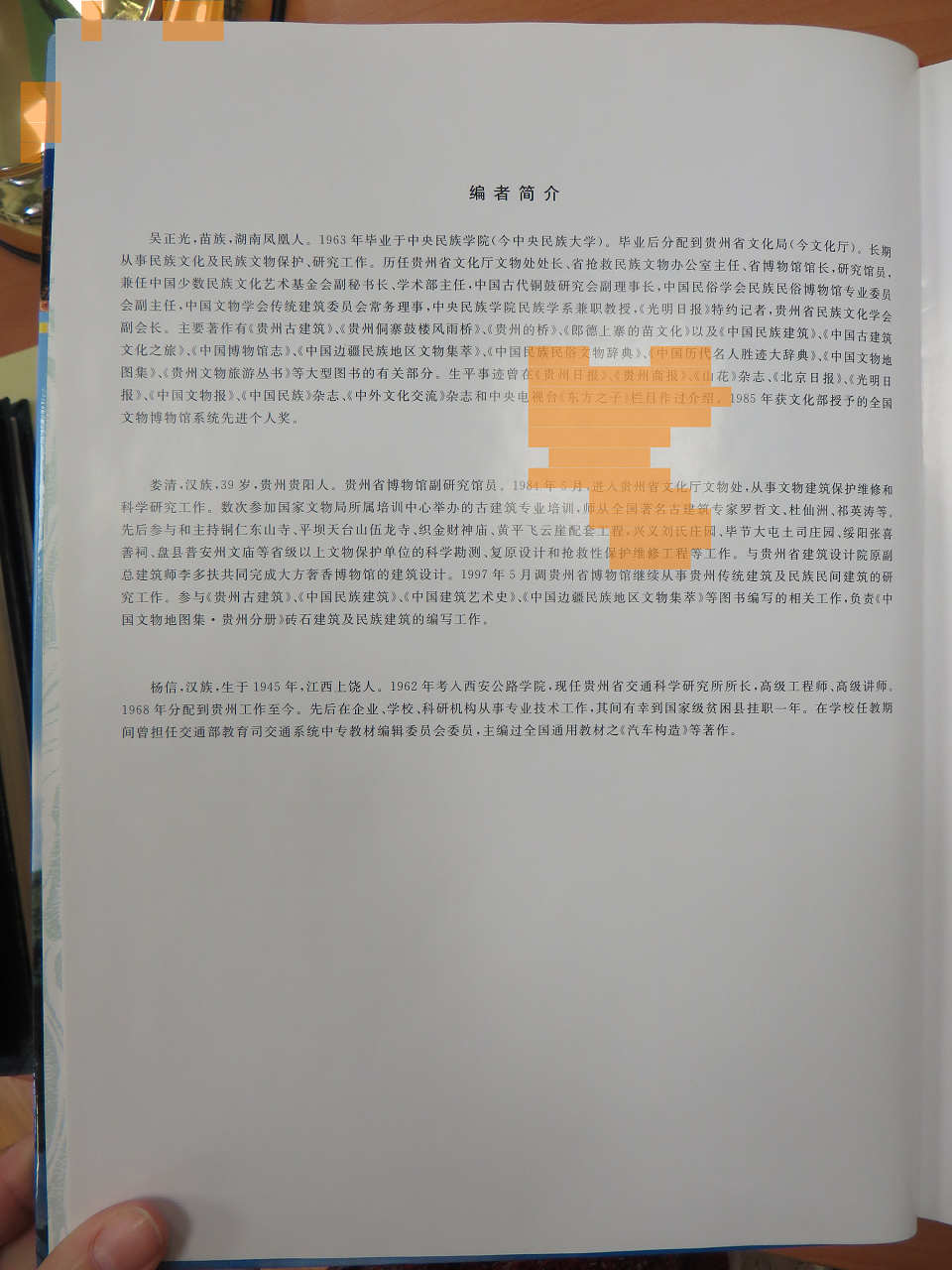}\\

\end{tabular}
\caption{Left-top: original image, right-top: image mark-up, left-bottom: mark-up to blocks conversion, right-bottom: predicted blocks with glare }
\end{figure}

This section provides information about the data used in the process of training and details about data preparation.

\begin{figure}
\includegraphics[width = 8.5cm, height=6cm]{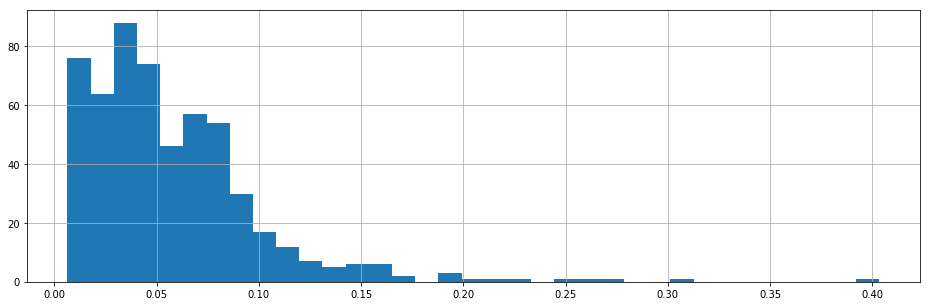}\\
\caption{Glare Area Distribution in dataset images}
\end{figure}

The model was trained on two datasets. Both datasets contained the same set of images. Each dataset consists of 692 images, with resolutions from full HD to 4k, made with mobile devices. Document types distribution in the dataset: 380 images of magazines, 198 documents images, and 114 business card images (all in Full HD resolution). The glare was present in all of the images. All of the images have normal luminance and the document is covering at least 60\% of the image. The average glare area is 5.8\% of the image, but it can cover up to 40\% of the image. The glare area distribution can be found in Fig. 2. Also, there is a difference in glare markup for datasets. For the first dataset, only glare in the document part of the image was labeled, for the second all existing glare were marked-up. In fact, markup consists of several bounding boxes. The example of an image, markup, converted to blocks markup and network prediction can be found in Fig. 1.

First, as augmentation, each image was resized with scale coefficient between $0.3$ and $1.5$.
Second, images were separated into square blocks of equal size. After that luminance features, such as minimal luminance, maximum luminance, dynamic range, mean luminance, and luminance deviation were collected. Third, the blocks were binarized using adaptive local binarization and the histograms of the black and white horizontal and vertical strokes length were built. To build the histograms we calculate the lengths of continuous runs of pixels of the same color in rows and columns.  The collected features from blocks were organized into matrices, with the first two dimensions corresponding to the relative position of the block in the image. By the presented preparations we obtain five matrices with features extracted from the original image. In the experiment, we used a block size of 64 pixels. As a result, from one image we obtain five different tensors, four for strokes histograms and one for luminance features, that we train our network on. 

To obtain the block labels from the original mark-up we intersect each block of the image with the mark-up region. The block is labeled as containing glare if the area of the intersection has at least $25\%$ of the block area. This also allows us to compensate for the original mark-up inaccuracy. The block in target heatmap was assigned to 1 if the corresponding block of the image intersected the markup as described. Otherwise, it was assigned to 0.

\section{Network Structure}
This section provides the details on the presented approach to the glare detection problem. Table 1 describes the feature extractor block of the network. Table 2 describes the predictor block of the network that generates the glare heat map based on the features obtained from the predictor block.

In this experiment, we focused on a lightweight network that will be capable to detect glare region on a mobile device at close to the real-time speed. We, therefore, focus on small models with straightforward architecture.

The network consists of five feature extraction blocks and a predictor block, therefore the network has five inputs. The feature extraction blocks take as an input a matrix of a histogram or luminance features. This input is then normalized using instance normalization [4] without any trainable parameters. This type of normalization allows us to ignore the luminance distribution in the given image and to concentrate more on the features, that are independent of the circumstances of when the image was taken and possessing more information about the presence of glare in the block of the image. The main idea of this block is to aggregate information from the original data based on the context of the block. The outputs of all of the feature extraction blocks are then concatenated and provided to the predictor block as an input.

The predictor block takes as an input a concatenation of the outputs of the feature extraction blocks. The structures of each block are described in Table I and Table II. It consists of 4 convolutional layers, that increase the theoretical receptive field of the model and allow to use more context information in glare detection. In theory, the average glare region on the image in 4K resolution should be covered by the receptive field.

\begin{table}[!t]
\renewcommand{\arraystretch}{1.3}
\caption{Feature extraction block}
\label{table_1}
\centering
\begin{tabular}{|c||c||c|}
\hline 
Layer & Kernel size & Theoretical receptive field \\
\hline 
Instance Normalization & $1 \times 1$ & $1 \times 1$  \\ 
\hline 
Convolutional & $1 \times 1$ & $1 \times 1$ \\ 
\hline 
Convolutional & $3 \times 3$ & $3 \times 3$ \\ 
\hline 
Convolutional & $3 \times 3$ & $5 \times 5$ \\ 
\hline 
\end{tabular} 
\end{table}

\begin{table}[!t]
\renewcommand{\arraystretch}{1.3}
\caption{Predictor block}

\label{table_2}
\centering
\begin{tabular}{|c||c||c|}
\hline 
Layer & Kernel size & Theoretical receptive field \\
\hline 
Convolutional & $3 \times 3$ & $7 \times 7$  \\ 
\hline 
Convolutional & $3 \times 3$ & $9 \times 9$ \\ 
\hline 
Convolutional & $3 \times 3$ & $11 \times 11$ \\ 
\hline 
Convolutional & $1 \times 1$ & $11 \times 11$ \\ 
\hline 
\end{tabular} 
\end{table}

\begin{table}[!t]
\renewcommand{\arraystretch}{1.3}
\caption{Unet-like architecture}

\label{table_3}
\centering
\begin{tabular}{|c||c||c||c|}
\hline 
Layer & Kernel size & Channels & Input Layers \\
\hline 
Instance normalization (1) & $1 \times 1$ & 1 & -  \\ 
\hline 
Convolutional (2) & $3 \times 3$ & 32 & 1\\ 
\hline 
Max Pooling (3) & $2 \times 2$ & 32 & 2\\ 
\hline 
Convolutional (4)& $3 \times 3$ & 32 & 3 \\ 
\hline 
Max Pooling (5)& $2 \times 2$ & 32 & 4 \\ 
\hline 
Convolutional (6) & $3 \times 3$ & 64 &  5 \\ 
\hline 
Max Pooling (7)& $2 \times 2$ & 64 & 6 \\ 
\hline 
Convolutional (8) & $3 \times 3$ & 64 &  7 \\ 
\hline 
Max Pooling (9)& $2 \times 2$ & 64 & 8 \\ 
\hline 
Convolutional (10) & $3 \times 3$ & 64 &  9 \\ 
\hline 
Max Pooling (11)& $2 \times 2$ & 64 & 10 \\ 
\hline 
Convolutional (12) & $3 \times 3$ & 64 &  11 \\ 
\hline

Convolutional (13) & $3 \times 3$ & 128 &  12 \\ 
\hline 
Up Sampling (14) & $2 \times 2$ & 128 &  13 \\ 
\hline 
Convolutional (15) & $3 \times 3$ & 128 &  [14,9] \\ 
\hline 
Up Sampling (16) & $2 \times 2$ & 128 &  15 \\ 
\hline 
Convolutional (17) & $3 \times 3$ & 128 &  [16,7] \\ 
\hline 
Convolutional (18) & $3 \times 3$ & 1 &  17 \\ 
\hline 
\end{tabular} 
\end{table}

\section{Experimental results}

The main focus of our work is to detect the approximate position of glare in short time and to have possibility present the result to the user of the mobile device to prevent the user from making a photo that would be unrecognizable. Also getting perfect mark-up of glare in the image is a complex task, because of the large variety of shapes of glare regions. That is why we decided to use recall and F-measure as main metrics. 

In the experiments, we focused on maximizing the recall and F-measure whilst maintaining the number of weights in the network less than $500,000$ and the amount of time on forward pass less than $40 ms$. In our experiments we used Tensorflow [5] as framework and Adam [6] optimized with a learning rate of  0.001, $\beta_1 = 0.99$, $\beta_2 = 0.999$. For the loss function, we used weighted cross entropy.

\subsection{Network baseline and comparison}

We decided to use naive Bayes classifier trained on luminance features described in section II. The baseline f-measure is $0.557$
and baseline recall is $0.502$. For further comparison, we decided to train an Unet-like[7] network, described in Table III, and compared inference time, number of weights and achieved f-measures. We trained the network on gray images and labeled groups of pixels sized $8 \times 8$.

\subsection{Network Training}

Before training, all images were rescaled as described in section II and the luminance features and strokes histograms were collected.

The network was first trained on the first dataset using random batches for 1500 epochs. Then we proceeded to train on hard-negative batches with the same dataset for 250 epochs. Finally, as the main focus is detecting glare on a document, we switched to the second dataset, where glare was marked-up only on document part of the image, and trained with hard-negative batches for 250 epochs. In each batch, we used random cropping to select patches of equal dimensions from images. To obtain a hard-negative batch we evaluated the model on the training part of the dataset and selected images with the highest loss.

\subsection{Results}

\begin{table}[!t]
\renewcommand{\arraystretch}{1.3}
\caption{Results}

\label{table_4}
\centering
\begin{tabular}{|c||c||c||c|}
\hline 
Network & F-measure & Total run time, ms & Size \\
\hline 
U-net & 0.798 & 5350 &  323,489 \\ 
\hline 
Our network & 0.740 & 151 & 405,473\\ 
\hline 

\end{tabular} 
\caption{By total run time we mean network inference time with feature collection and preprocessing, if needed. }
\end{table}

To calculate precision and recall we output matrix of the network to binary matrix by assigning every block with a value higher than threshold a 1, and all of the others a zero. The resulting time needed for the forward pass is $31 ms$, the time for all of the feature collection is $120ms$ and the total number of weights of $405,473$. The obtained metrics can be found in Fig. 3.
The maximum F-measure is 0.740, with the recall of 0.736 and precision of 0.744, achieved at glare in block threshold of 0.9. The comparison with the other network can be found in Table IV. All the measurements were performed on
i5-7500 3.40 GHz CPU and 16 GB RAM on images in 4K resolution.

\begin{figure}[!t]
    \centering
    \includegraphics[width=8.3cm, height=12cm]{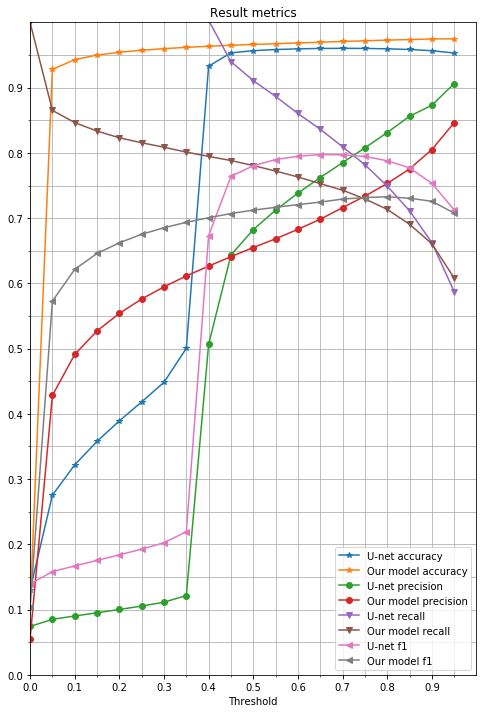}
    \caption{Experimental results}
    \label{fig:expreiment_results}
\end{figure}

\section{Conclusion}

We have presented a novel method for glare detection, which is able to detect glare and could be launched on mobile devices. The presented network has shown to be much faster than U-net like architectures. The core idea of the model is based on special handcrafted features extracted from the document, which reduces the complexity of the model and provides high F-measure.



%

\end{document}